\documentclass[english,runningheads,a4paper]{llncs}
\usepackage{graphicx}
\usepackage{subcaption}
\usepackage{url}
\usepackage{multirow}
\usepackage{float}
\usepackage{tabularx}
\usepackage{makecell}
\usepackage{tabulary}
\usepackage{hyperref}
\usepackage{xcolor}
\usepackage{hhline}
\usepackage{array}
\usepackage{amsfonts}
\usepackage{amsmath}
\usepackage{fancyhdr}
\usepackage{soul}
\usepackage{pgffor}
\usepackage{xparse}
\usepackage{ifthen}
\usepackage{tikz}
\usepackage{color}

\usepackage{makecell}
\usepackage[symbol]{footmisc}

\newcolumntype{P}[1]{>{\centering\arraybackslash}p{#1}}

\DeclareDocumentEnvironment{env_comparison_0}{O {10} m}
{ 
\begin{figure}[htb]
	\centering
	\begin{subfigure}[t]{\textwidth}
		\centering
		\includegraphics[width=1.17in]{figures/suppl/student_default/#1_dt_2d.png}	
		\includegraphics[width=1.17in]{figures/suppl/student_default/#1_dt_3d.png}	
		\hfill	
		\includegraphics[width=1.17in]{figures/suppl/student_default_12x/#1_dt_2d.png}
		\includegraphics[width=1.17in]{figures/suppl/student_default_12x/#1_dt_3d.png}			
		\caption{Default student network's results for 1x and 12x downsampling scale}
	\end{subfigure}	
	\begin{subfigure}[t]{\textwidth}
		\centering
		\includegraphics[width=1.17in]{figures/suppl/orpose_all_1x/#1_dt_2d.png}
		\includegraphics[width=1.17in]{figures/suppl/orpose_all_1x/#1_dt_3d.png}	
		\hfill
		\includegraphics[width=1.17in]{figures/suppl/orpose_all_12x/#1_dt_2d.png}		
		\includegraphics[width=1.17in]{figures/suppl/orpose_all_12x/#1_dt_3d.png}			
		\caption{Trained student network's (ORPose\_all) results for 1x and 12x downsampling scale}
	\end{subfigure}	
	\label{figure:qual-results-1}
\end{figure}
}
{}

\DeclareDocumentEnvironment{env_comparison_1}{O {10} m}
{   
  \begin{figure}[!htb]
    \centering 
    \begin{subfigure}[b]{.32\linewidth}
      \includegraphics[width=\linewidth]{figures/suppl/gt/#1_gt_2d.jpg}
      \caption{GT 2D}
    \end{subfigure}
    \begin{subfigure}[b]{.32\linewidth}
      \includegraphics[width=\linewidth]{figures/suppl/orpose_fixed_1x/#1_dt_2d.jpg}
      \caption{ORPose\_fixed\_1x 2D}
    \end{subfigure}     
    \begin{subfigure}[b]{.32\linewidth}
      \includegraphics[width=\linewidth]{figures/suppl/orpose_fixed_12x/#1_dt_2d.jpg}    
      \caption{ORPose\_fixed\_12x 2D}
    \end{subfigure} \\
    \begin{subfigure}[b]{.32\linewidth}
      \includegraphics[width=\linewidth]{figures/suppl/gt/#1_gt_3d.jpg}
      \caption{GT 3D}
    \end{subfigure}
    \begin{subfigure}[b]{.32\linewidth}
      \includegraphics[width=\linewidth]{figures/suppl/orpose_fixed_1x/#1_dt_3d.jpg}
      \caption{ORPose\_fixed\_1x 3D}
    \end{subfigure}
    \begin{subfigure}[b]{.32\linewidth}
      \includegraphics[width=\linewidth]{figures/suppl/orpose_fixed_12x/#1_dt_3d.jpg}    
      \caption{ORPose\_fixed\_12x 3D}
    \end{subfigure} \\
    \label{fig}
    \end{figure}
}
{}

\DeclareDocumentEnvironment{env_comparison_2}{O {10} m}
{   
  \begin{figure}[!htb]
    \centering 
    \begin{subfigure}[b]{.32\linewidth}
      \includegraphics[width=\linewidth]{figures/suppl/gt/#1_gt_2d.jpg}
      \caption{GT 2D}
    \end{subfigure}
    \begin{subfigure}[b]{.32\linewidth}
      \includegraphics[width=\linewidth]{figures/suppl/orpose_all_1x/#1_dt_2d.jpg}
      \caption{ORPose\_all\_1x 2D}
    \end{subfigure}   
    \begin{subfigure}[b]{.32\linewidth}
      \includegraphics[width=\linewidth]{figures/suppl/orpose_all_12x/#1_dt_2d.jpg}    
      \caption{ORPose\_all\_12x 2D}
    \end{subfigure} \\
    \begin{subfigure}[b]{.32\linewidth}
      \includegraphics[width=\linewidth]{figures/suppl/gt/#1_gt_3d.jpg}
      \caption{GT 3D}
    \end{subfigure}
    \begin{subfigure}[b]{.32\linewidth}
      \includegraphics[width=\linewidth]{figures/suppl/orpose_all_1x/#1_dt_3d.jpg}
      \caption{ORPose\_all\_1x 3D}
    \end{subfigure}  
    \begin{subfigure}[b]{.32\linewidth}
      \includegraphics[width=\linewidth]{figures/suppl/orpose_all_12x/#1_dt_3d.jpg}    
      \caption{ORPose\_all\_12x 3D}
    \end{subfigure} \\
    \label{fig}
    \end{figure}
}
{}

\begin{document}
\title{Self-supervision on Unlabelled OR Data for Multi-person 2D/3D Human Pose Estimation}
% \title{Knowledge Distillation from Unlabelled Data: A self-supervised Approach for Multi-person 2D/3D Human Pose Estimation}
%\titlerunning{Confidential submission - Paper ID \# 2324}

%\author{*** *** *** }

%\authorrunning{** ** **}

% \author{Vinkle Srivastav\inst{1}\orcidID{https://orcid.org/0000-0002-1044-038X} \and Afshin Gangi\inst{1,2}\orcidID{https://orcid.org/0000-0002-0952-7760} \and Nicolas Padoy\inst{1}\orcidID{https://orcid.org/0000-0002-5010-4137}}
\author{Vinkle Srivastav\inst{1}\and Afshin Gangi\inst{1,2}\and Nicolas Padoy\inst{1}}
% index{Srivastav, Vinkle}
% index{Gangi, Afshin}
% index{Padoy, Nicolas}

%\authorrunning{Confidential submission - Paper ID \# 2324}

% First names are abbreviated in the running head.
% If there are more than two authors, 'et al.' is used.
%

\urldef{\mailsa}\path{{srivastav | padoy}@unistra.fr}
 \institute{ICube, University of Strasbourg, CNRS, IHU Strasbourg, France\\ \mailsa
   \and Radiology Department, University Hospital of Strasbourg, France}
 
%\institute{***}
%
\maketitle              % typeset the header of the contribution
\begin{abstract}
	2D/3D human pose estimation is needed to develop novel intelligent tools for the operating room that can analyze and support the clinical activities. The lack of annotated data and the complexity of state-of-the-art pose estimation approaches limit, however, the deployment of such techniques inside the OR. In this work, we propose to use knowledge distillation in a teacher/student framework to harness the knowledge present in a large-scale non-annotated dataset and in an accurate but complex multi-stage teacher network to train a lightweight network for joint 2D/3D pose estimation. The teacher network also exploits the unlabeled data to generate both hard and soft labels useful in improving the student predictions. The easily deployable network trained using this effective self-supervision strategy performs on par with the teacher network on \emph{MVOR+}, an extension of the public MVOR dataset where all persons have been fully annotated, thus providing a viable solution for real-time 2D/3D human pose estimation in the OR.
\end{abstract}

\footnotetext{Accepted at International Conference on Medical Image Computing and Computer-Assisted Intervention - MICCAI 2020.}
\thispagestyle{fancy}
\pagestyle{fancy}
\fancyhf{}
\renewcommand{\headrulewidth}{0pt}
%\fancyfoot[C]{{ *** Confidential submission - Paper ID \#2324 ***}}

\section{Introduction}
The current surge in deep learning research has made its way into the operating room (OR) with the goal to develop novel intelligent context-aware assistance systems \cite{maier2017surgical,vercauteren2019cai4cai}. Human pose estimation (HPE), a key requirement to build such systems, is a computer vision task aiming at localizing human body parts, also called keypoints, either in 2D or 3D. Owing to the current advancements in deep learning, HPE has made remarkable progress and works reliably on \textit{in the wild} images. These advances can be attributed to the emergence of large scale supervised datasets, such as COCO \cite{lin2014microsoft} and Human3.6 \cite{h36m_pami}, and to the modern deep learning architectures. These architectures employ multi-stage deep neural networks to achieve state-of-the-art performance. For example, top approaches for the task of 2D pose estimation on the COCO dataset use a two-stage approach, in which the first stage determines the person's bounding boxes and the second stage estimates the keypoints for each bounding box. This multi-stage design using high-capacity deep neural networks in both stages helps to achieve better accuracy. The run-time performance of such a design is however considerably low. More practical solutions such as Mask-RCNN \cite{he2017mask} or OpenPose \cite{cao2016realtime} use low-capacity and single-stage networks for the same task to achieve better run-time performance, but the accuracy of these system is low compared to the multi-stage systems. Therefore, one key challenge for the deployment of HPE network inside the OR is not only to give accurate predictions, but also to be fast using a light-weight and single-stage end-to-end design, as needed for real-time applications. 

Another key challenge is the decrease in the accuracy of state-of-the art approaches when applied to a different target domain, such as the OR \cite{srivastav2018mvor}. This decrease can be compensated by finetuning the approach on manual annotations from the target domain, as done by \cite{kadkhodamohammadi2017articulated,kadkhodamohammadi2017multi,belagiannis2016parsing,hansen2019fusing} for the OR. Obtaining manually annotated data is however time-consuming and expensive. Specifically, in the case of the operating room, utilizing crowd sourcing technique such as Amazon Mechanical turk would be impractical due to  obvious privacy concerns. Therefore, techniques that can harness knowledge from non-annotated datasets captured in the target environment are extremely useful.

In this paper, our work lies at the intersection of knowledge distillation \cite{hinton2015distilling,zhang2019fast} and data distillation \cite{radosavovic2018data} and exploits these techniques to solve the task of multi-person 2D/3D HPE. We use knowledge distillation to transfer knowledge from an accurate, larger, and multi-stage teacher network to a practical, smaller, and single-stage student network. The idea of knowledge distillation has been adapted to different problems. In \cite{hinton2015distilling}, authors use the probability output vector of a teacher network as soft-labels to train a student network for multi-class classification. The student network learns jointly from the soft-labels generated by the teacher output and from the hard-labels given by the ground truth. Similarly, in \cite{zhang2019fast}, authors use this approach for single-person 2D HPE by jointly training the student network from the soft output heatmaps of a teacher network and the hard ground truth heatmaps. Both the student and the teacher network work on the fully supervised dataset, and the soft output of the teacher network serves as additional useful labels along with the hard labels obtained from the supervised ground truth. 

In this work, we aim at applying knowledge distillation when only {\it non-annotated} data is available in the target domain. Instead of using supervised annotations, we propose to use data distillation to generate labels automatically from the non-annotated dataset. We run a complex teacher network which ensembles output predictions on geometrically transformed input images. Unlike the standard use of data distillation \cite{radosavovic2018data}, which only exploits hard predictions obtained by removing the low confidence keypoints, we also use the soft predictions from the confidence value for each keypoint. 

%We generate two prediction sets from the teacher network on the non-annotated data: a \textit{hard-set} containing high confidence scores, and a \textit{soft-set} containing confidence values along with the keypoints coordinates. We argue that these two sets provide useful signal from the non-annotated data.

As student network, we use a low capacity single stage network based on Mask-RCNN. The architecture of the student network is inspired from \cite{dabral2019multi} and further extended to effectively use the \textit{hard-set} and \textit{soft-set} for joint 2D/3D multi-person HPE in the OR. By utilizing our approach, the student network reaches an accuracy on par with the teacher network. 

%For 2D Keypoint estimation task on the average precision (AP) metric, our student network improves from 46.17 AP to 55.80 AP while reaching close to the teacher network having 57.78 AP; for the 3D Keypoint estimation task on the mean per joint position error (MPJPE) metric, error in our student network decreases from 147.17 mm to 134.13 mm surpassing the teacher network by 0.75 mm. Existing approaches for the HPE inside the OR work in a fully supervised fashion \cite{kadkhodamohammadi2017articulated, kadkhodamohammadi2017multi,belagiannis2016parsing}; we differ from these approaches based on our novel formulation to exploit the non-annotated data. 

Another specific issue in the OR is to preserve the privacy of patients and clinicians while performing computer vision tasks. Human pose estimation on low-resolution images has been suggested to improve the privacy \cite{srivastav2019human}, like for activity recognition \cite{haque2018activity}. We therefore also extend our approach to deliver accurate poses on low-resolution images. 
%We propose a simple approach to adapt our student network so that it performs considerably well on low-resolution images. During training, we train the network on high resolution as well as on extremely low-resolution images using a downsampling factor up to 12x. 
This turns out to be very effective on the OR images compared to the teacher network, even with a 12x downsampling factor. 

\section{Methodology}
\subsection{Problem overview}
\begin{figure}[tb]
	\includegraphics[width=1.0\linewidth]{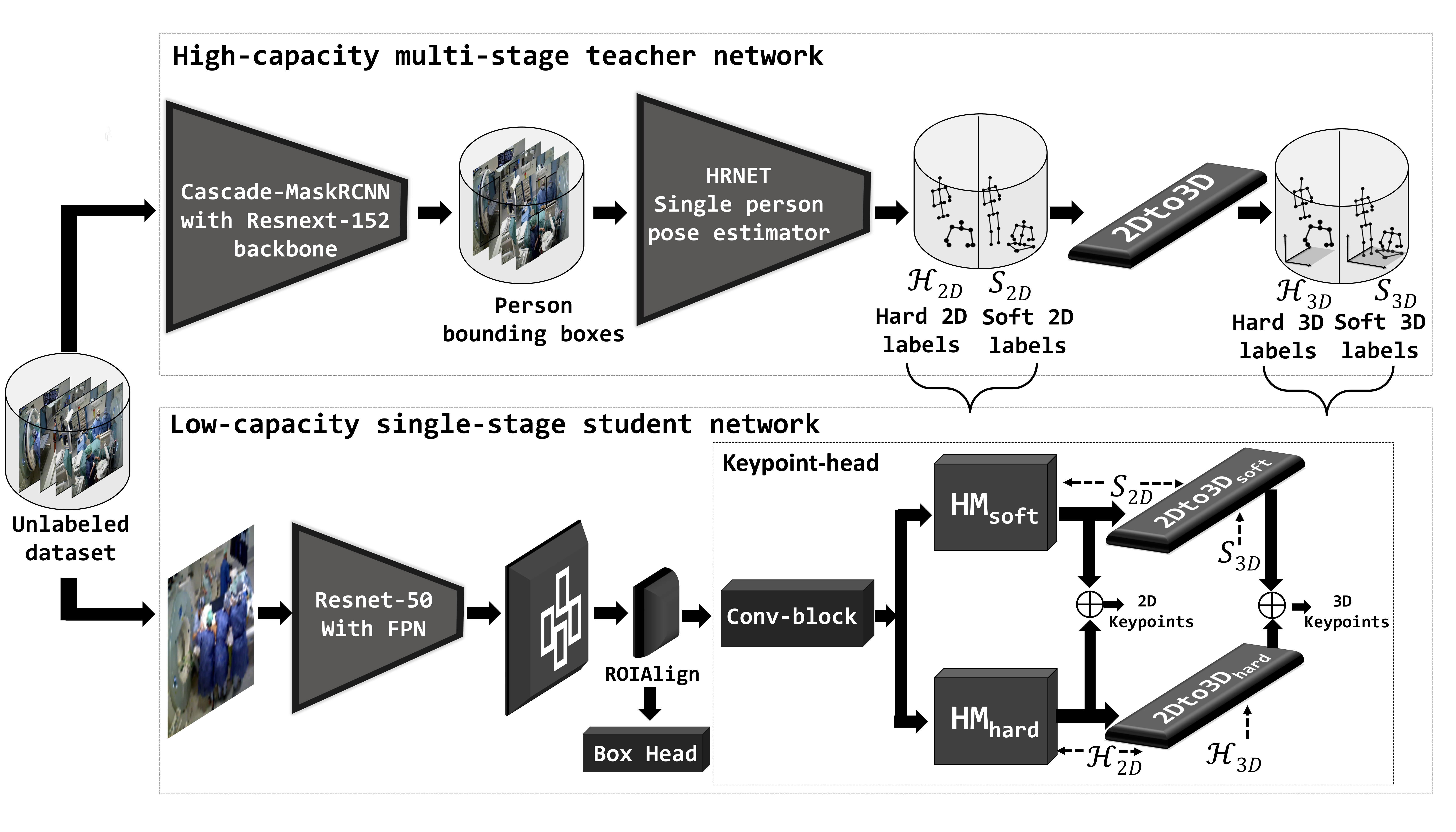}
	\caption{ Proposed self-supervised methodology for joint 2D/3D keypoint estimation using the teacher/student paradigm. The teacher network is a three-stage network which uses the unlabelled dataset to extract person bounding boxes, estimate 2D keypoints, and regress 2D keypoints to 3D. It generates soft and hard pseudo labels to be used by the student network. The student network is a single-stage network and effectively utilizes the soft and hard pseudo labels to jointly estimate the 2D and 3D keypoints.}
	\label{fig:architecture}
\end{figure}
Given a monocular RGB input image $\mathcal{I}$ of size $W \times H$, our task is to detect the 2D and 3D body keypoints for multiple persons using a single efficient end-to-end network. The 2D keypoints $\mathcal{P}_{2D} \in \mathbb{R}^{m \times n \times 2} $ are in image coordinates, and the 3D keypoints $\mathcal{P}_{3D} \in \mathbb{R}^{m \times n \times 3} $ are in the root-relative coordinates (where the root-joint is set to be the origin and all other joints are measured w.r.t root-joint). Here, $m$ is the number of persons, and $n=17$  is the number of joints for each pose. We also consider low-resolution images for the same task, which have very small input sizes. To tackle the problem, we utilize a teacher/student approach to train the end-to-end student network by distilling the knowledge in a teacher network on large-scale unlabelled data. We follow a two-step approach: in the first step, a multi-stage high-capacity neural network (a teacher network) is used to generate pseudo labels; in the second step, these pseudo labels are used to train an end-to-end low-capacity network (a student network).
\subsection{Knowledge generation using the teacher network}
The teacher network, shown in Figure \ref{fig:architecture}, is a three-stage network: The first stage uses the cascade-mask-rcnn \cite{cai2019cascade} with the resnext-152 \cite{xie2017aggregated} backbone to generate person bounding boxes, the second stage estimates the 2D keypoints for each bounding box using the HRNet architecture \cite{SunXLW19} after discarding low-score bounding boxes, and the third stage lifts the detected 2D keypoints to the 3D using a residual-based 2-layer fully-connected network \cite{martinez2017simple}. The three stages in the teacher network are selected based on their state-of-art performance on the COCO and Human3.6 dataset.
The first and second stages are trained on the COCO dataset \cite{lin2014microsoft} and the third stage is trained on the Human3.6 dataset \cite{h36m_pami}. Multi-level scaling and flipping transformations are applied in the first and the second stage to obtain good quality person bounding boxes and 2D keypoints. However, errors can still be present in the keypoints and are encoded in the keypoint confidence scores. Therefore, we construct two sets of pseudo-labels: the soft-set $\mathcal{S}$ and the hard-set $\mathcal{H}$. The soft-set $ \mathcal{S} = \{\mathcal{S}_{2D}, \mathcal{S}_{3D} \}$ consists of soft 2D keypoints and soft 3D keypoints. Soft 2D keypoints $\mathcal{S}_{2D} \in \mathbb{R}^{m \times n \times 3}$ are obtained by storing the confidence value for each keypoint along with their coordinates. The last dimension in $\mathbb{R}^{m \times n \times 3}$ represents the channel for the confidence value. $\mathcal{S}_{2D}$ is sent to the third stage to obtain the soft 3D keypoints $\mathcal{S}_{3D}$.
Similarly, the hard-set $ \mathcal{H} = \{\mathcal{H}_{2D}, \mathcal{H}_{3D} \}$ consists of hard 2D keypoints and hard 3D keypoints. $\mathcal{H}_{2D}$ is obtained by only keeping the high confidence 2D keypoints and discarding the low confidence keypoints. $\mathcal{H}_{3D}$ is obtained by passing the $\mathcal{H}_{2D}$ to the lifting network. We show in the experiments that these two sets provide useful learning signals when used to train the student. In the next section, we show how we exploit these two sets of pseudo labels for effectively training the student network.

% $$ \mathcal{S} = \{P_{2dsoft}, {P_{3dsoft}}\} ; P_{2dsoft} \in \mathbb{R}^{m \times n \times 3}, P_{3dsoft} \in \mathbb{R}^{m \times n \times 3}$$
% $$ \mathcal{H} = \{P_{2dhard}, {P_{3dhard}}\}; P_{2dhard} \in \mathbb{R}^{m \times n \times 2}, P_{3dhard} \in \mathbb{R}^{m \times n \times 3} $$

\subsection{Knowledge distillation in the student network:}
% assume we H2d, H3d, S2d, S3d
The student network presented in Figure \ref{fig:architecture} is an end-to-end network based on Mask-RCNN that jointly predicts the 2D and 3D poses. We replace the mask head of the Mask-RCNN network with a keypoint-head for joint 2D and 3D pose estimation. The keypoint-head accepts the fixed size proposals from the ROIAlign layer and passes them through 8 conv-block layers to generate the features. These features are upsampled using a deconv and bi-linear upsampling layer into two branches to generate 17 channel heatmaps corresponding to each body joint. The first branch upsamples the features to generate the heatmaps $HM_{soft}$, and the second branch upsamples them to generate the heatmaps $HM_{hard}$. The $HM_{soft}$ and $HM_{hard}$ heatmaps are connected to their respective lifting networks i.e $2Dto3D_{soft}$ and $2Dto3D_{hard}$ to lift the incoming 2D keypoints to 3D.

\noindent\textbf{Training}:
Training of the network follows the same framework as Mask-RCNN along with the additional losses coming from the keypoint-head. In the keypoint-head, we compute 2D and 3D losses $L_{2D}$ and $L_{3D}$ to estimate the 2D and 3D keypoints. $L_{2D}$ consists of soft and hard 2D keypoint losses. The soft 2D keypoint loss $L_{2Dsoft}$ is obtained by first multiplying  $HM_{soft}$ with the corresponding confidence values from the last channel of $\mathcal{S}_{2D}$ and then computing its cross-entropy loss with $\mathcal{S}_{2D}$. The hard 2D keypoint loss $L_{2Dhard}$ is obtained by calculating the cross-entropy loss between $HM_{hard}$ and $\mathcal{H}_{2D}$. Similarly, the 3D loss $L_{3D}$ consists of soft and hard 3D keypoint losses. Soft 3D keypoint loss $L_{3Dsoft}$ is obtained by taking the smooth L1 loss between $\mathcal{S}_{3D}$ and the output of $2Dto3D_{soft}$ using the input $\mathcal{S}_{2D}$, and hard 3D keypoint loss $L_{3Dhard}$ is obtained by taking the smooth L1 loss between $\mathcal{S}_{3D}$ and the output of $2Dto3D_{hard}$ using the input $\mathcal{H}_{2D}$. All four losses are added together to obtain the loss for the keypoint-head $L_{kpt}$. The overall loss is the sum of $L_{kpt}$ with the standard Faster-RCNN loss, ie. the bounding box classification and regression loss, and the region proposal loss.

\noindent\textbf{Inference}:
During inference, the 2D keypoints are computed by taking the arg-max over each channel from the mean output of $HM_{soft}$ and $HM_{hard}$, and the 3D keypoints are computed by calculating the 2D keypoints from $HM_{soft}$ and $HM_{hard}$ using arg-max, passing the 2D keypoints to the respective 2Dto3D lifting network, and averaging the 3D output.

\begin{figure}[t]
	%\hfill
	\centering
	\begin{subfigure}[t]{1.5in}
		\centering
		\includegraphics[width=1.5in]{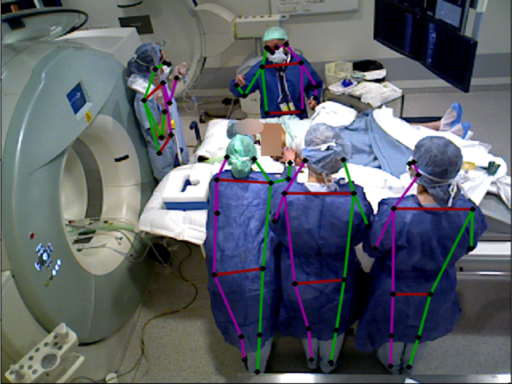}
		\caption{GT-2D}
	\end{subfigure}	
	\begin{subfigure}[t]{1.5in}
		\centering
		\includegraphics[width=1.5in]{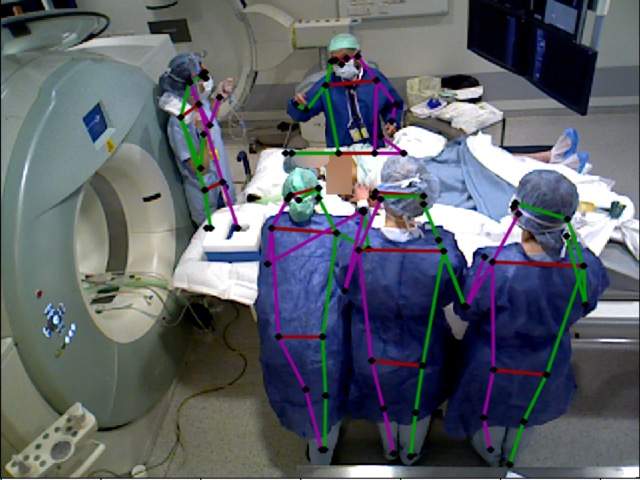}	
		\caption{1x (640x480)}
	\end{subfigure}	
	\begin{subfigure}[t]{1.5in}
		\centering
		\includegraphics[width=1.5in]{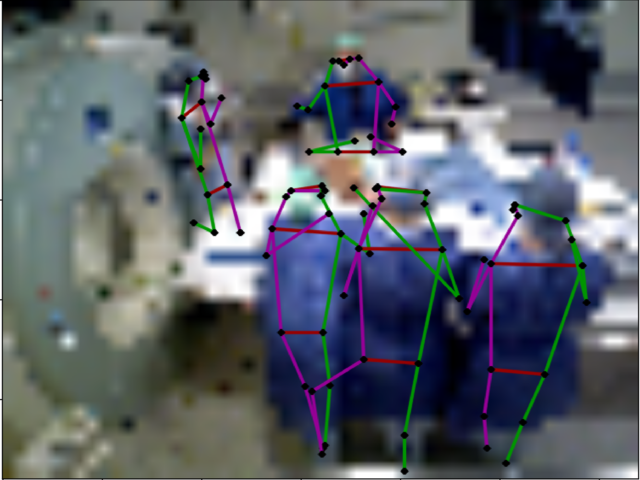}
		\caption{12x (53x40)}
	\end{subfigure}		
	\begin{subfigure}[t]{1.5in}
		\centering
		\includegraphics[width=1.5in]{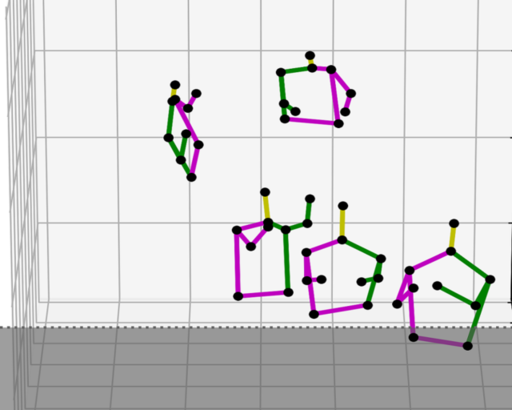}
		\caption{GT-3D}
	\end{subfigure}	
	\begin{subfigure}[t]{1.5in}
		\centering
		\includegraphics[width=1.5in]{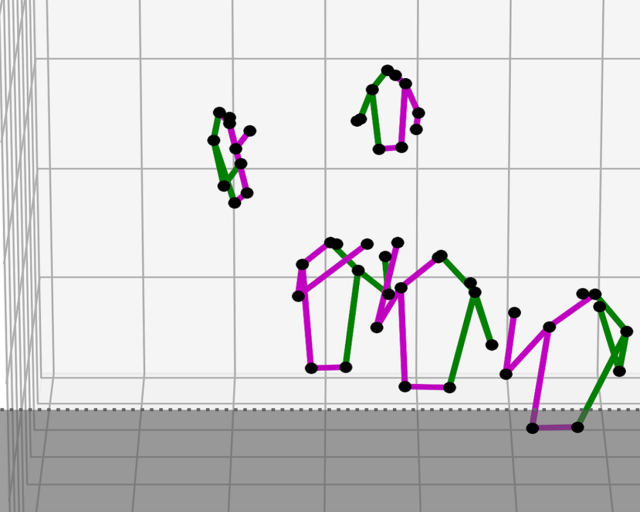}
		\caption{1x (640x480)}
	\end{subfigure}	
	\begin{subfigure}[t]{1.5in}
		\centering
		\includegraphics[width=1.5in]{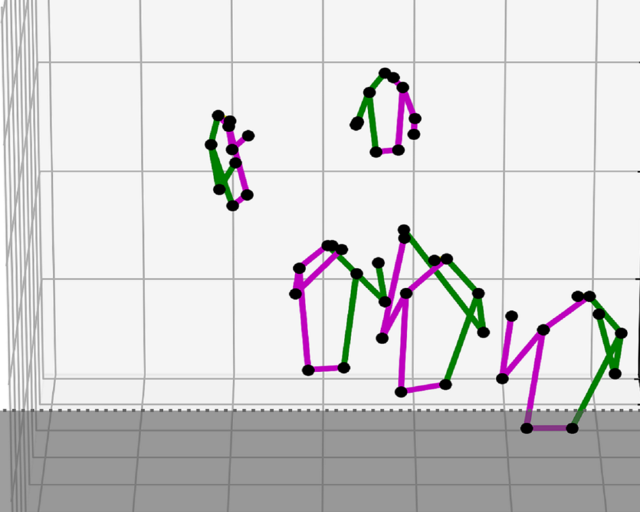}
		\caption{12x (53x40)}
	\end{subfigure}		
	\caption{Qualitative results for 2D and 3D keypoints estimation from the student network (ORPose\_all) at original and downsampled image sizes. GT-2D and GT-3D are the visualization results from 2D and 3D ground truth keypoints respectively. Since we are not predicting the scale of the 3D pose in the camera frame, we use the depth of the root node from the ground truth as scale to generate this visualization.}
	\label{result_1}
\end{figure}

% $$ \mathcal{L}_{2dhard}(\mathcal{I}) = \frac{1}{m \times v} cross\_entropy(HM_{hard}, P_{2dhard})$$
% $$ \mathcal{L}_{2dsoft}(\mathcal{I}) = \frac{1}{m \times v} cross\_entropy(HM_{soft} \times P_{2dsoft}(:,:,3) , P_{2dsoft}(:,:,:2))$$

% \subsubsection{Root-joint Optimization} % needs to be done
% We use the weak-perspective model to project each 3D pose into the camera coordinates. It is just an approximation which places the persons appropriately in the depth based on their scale. we chose to include only the upper body in the formula because most of the time lower body joints are occluded. explain the formula

\section{Experiments and Results}
% The 2D network is pretrained on the coco-dataset whereas the lifting network is pre-trained on the human3.6 dataset.
% The unlabelled dataset is equally distributed into four categories, in which the number of persons equal to one, two, three, or four or more.
\subsection{Training and Testing Dataset}

We use the public MVOR dataset \cite{srivastav2018mvor} as a test set. We also use an unlabeled dataset of 80k images to generate pseudo labels and train our networks. The unlabeled dataset consists of images captured on different days than MVOR to ensure that there is no overlap with the test set. We split the dataset into 77k train and 3k validation images. Note that the public MVOR dataset \cite{srivastav2018mvor} is not fully annotated: not all persons are annotated and 2D keypoints were only annotated for 10 upper body parts. To use this dataset in the standardized COCO evaluation framework for bounding boxes and 2D keypoints, we extend the dataset with additional annotations. Before the extension, MVOR consists of 4699 person bounding boxes, 2926 2D upper body poses with 10 keypoints, and 1061 3D upper body poses. The extended MVOR dataset, called MVOR+, consists of 5091 person bounding boxes and 5091 body poses with 17 keypoints in the COCO format. %1061 3D upper body poses are defined in MVOR for the multiview image (each multiview image consists of three images). 
%Since we are evaluating 3D poses for the single view, we projected these 1061 3D poses into the respective camera coordinates to obtain 2926 valid 3D poses (we discarded the not-visible poses). 
The original size of all the images is 640x480. We also conduct experiments with downsampled images using the scaling factors 8x, 10x, and 12x, yielding images of size 80x64, 64x48, and 53x40.

\subsection{Evaluation Metrics} We use the Average Precision (AP) metric from COCO to evaluate bounding box detection and 2D keypoint estimation. We use the 3D mean per joint position error (MPJPE) in millimeters to evaluate the 3D keypoints. The 2D keypoint AP is computed using 17 keypoints in COCO format whereas the MPJPE metric is computed using the 8 keypoint from upper-body pose (shoulder, elbow, hand, hip). High AP is desired for the bounding box detection and 2D keypoint estimation, while a low MPJPE score is desired for 3D keypoint estimation. For 3D evaluation, the 3D ground-truth keypoints are expressed in the camera coordinate frame and each joint in the 3D pose is subtracted from the pelvis root-joint (taken as the mean of left and right hips) to obtain the root-relative pose.
%The 3D ground truth in the MVOR dataset is defined for 10 upper body parts in the multiview setup i.e for each set of multiview images 3d poses are defined. Since we need to evaluate the pose single view we projected the images to the respective camera using the extrinsic and intrinsic camera parameters and the poses are subtracted from the pelvis joint (since pelvis joint doesn't exist in the MVOR, we calculated the pelvis location as the average of left and right hip joint).
% We use the detectron2 framework from the Facebook AI research group to train the student network. We extended the keypoint-rcnn model for our student network from the detectron2 framework which provides the flexibility using the registry mechanism to extend the model and to use the custom dataset. \\ $\mathcal{H}_{2D}$

\subsection{Experiments}

The student network is trained differently for two sets of experiments, yielding the networks ORPose\_fixed\_\textbf{s}x (\textbf{s}=1,8,10,12) and ORPose\_all. ORPose\_fixed\_\textbf{s}x is trained using either images of the original size (\textbf{s}=1) or low-resolution images at a fixed scaling factor (\textbf{s}=8,10,12). When feeding the networks, low-resolution images are first upsampled to match the original input size.\\
Evaluation of networks ORPose\_fixed\_\textbf{s}x is done on the same scale they are trained on. In the second experiment, ORPose\_all is trained using original and downsampled images with a random downsampling factor. This is similar to the scaling data augmentation technique, but we consider here a very low-resolution scenario where the input image is downsampled up to 12x. We choose the random scale such that for 30\% of the training time there is no downsampling, for 35\% the downsampling scale is randomly chosen between 2 and 8, and for the remaining 35\% of training time downsampling scale is randomly chosen between 8 and 12. The intention to train the network using this strategy is to obtain a single model that can work on high-resolution images and should also perform considerably better on the low-resolution images. 
The base learning rate for ORPose\_fixed\_1x is set to 1e-3 for 5k total number of iterations with a step decay of 0.1 after 2k, 3k, and 4k iterations; the base learning rate for ORPose\_fixed\_\textbf{s}x (\textbf{s}=8,10,12) is set to 1e-2 for 10k total number of iterations with a step decay of 0.1 after 7k, 8k, and 9k iterations; the base learning rate for ORPose\_all is set to 1e-1 for 20k total number of iterations with step decay of 0.1 after 14k, 16k, and 18k iterations. The downsampling and upsampling operation is performed using bilinear interpolation. We use a detectron2 framework \cite{wu2019detectron2} to run all the experiments on two V100 NVidia GPUs using the distributed data parallelism framework of PyTorch. We use a batch size of 32 and the stochastic gradient solver as the optimizer for all the experiments.

\begin{figure}[htb]
	%\hfill
	\centering
	\begin{subfigure}[t]{\textwidth}
		\centering
		\includegraphics[width=1.17in]{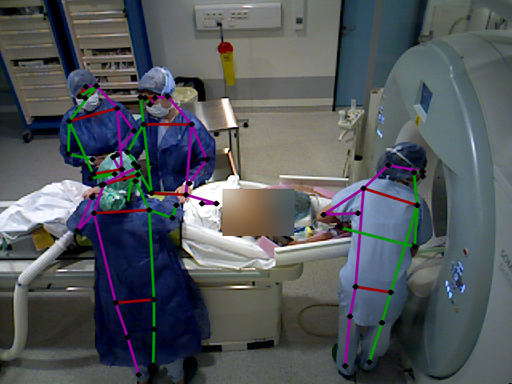}	
		\includegraphics[width=1.17in]{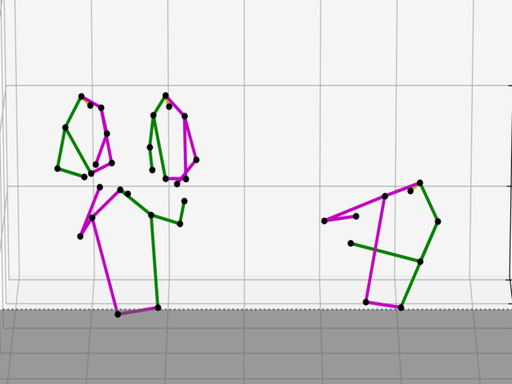}	
		\hfill	
		\includegraphics[width=1.17in]{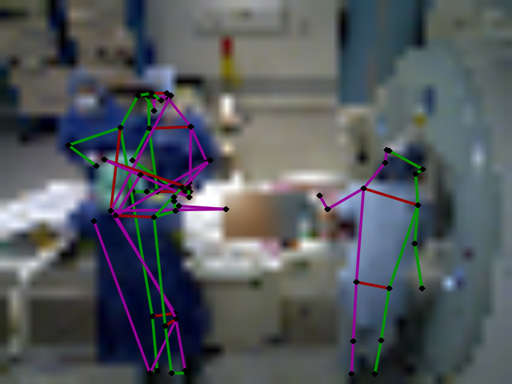}
		\includegraphics[width=1.17in]{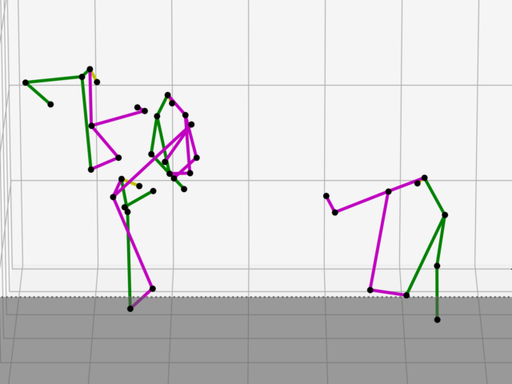}			
		\caption{Results of default student network for 1x and 12x downsampling scale}
	\end{subfigure}	
	\begin{subfigure}[t]{\textwidth}
		\centering
		\includegraphics[width=1.17in]{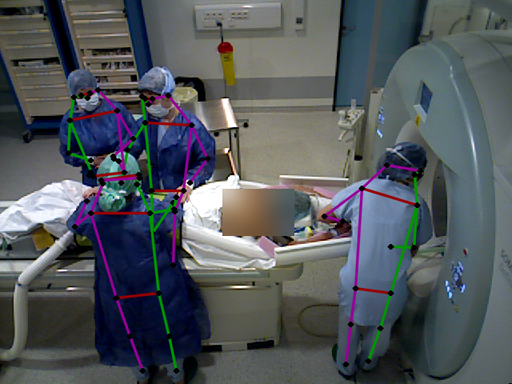}
		\includegraphics[width=1.17in]{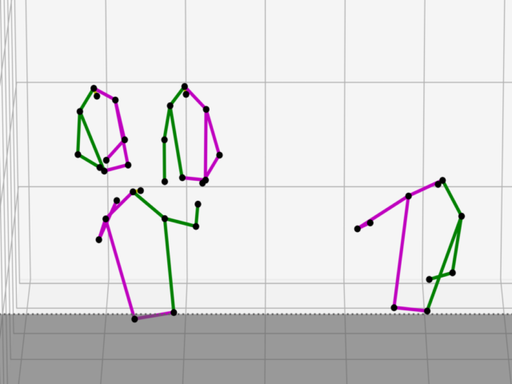}	
		\hfill
		\includegraphics[width=1.17in]{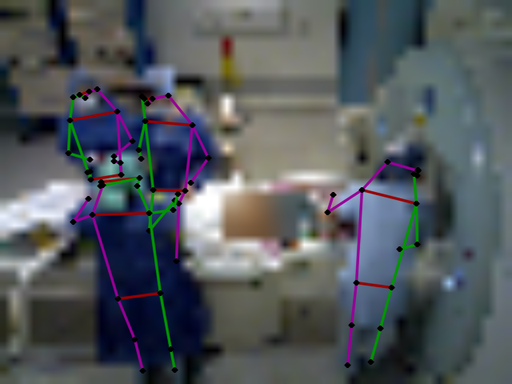}		
		\includegraphics[width=1.17in]{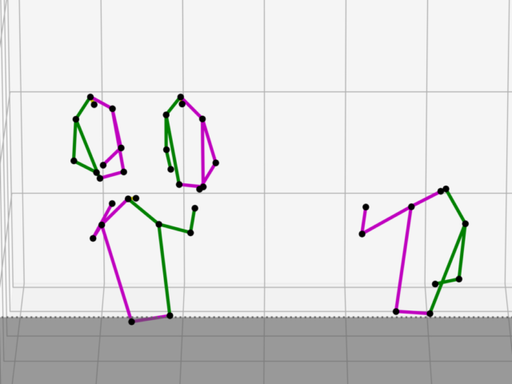}			
		\caption{Results of trained student network for 1x and 12x downsampling scale}
	\end{subfigure}	
	\caption{Comparative qualitative results for the default and the trained student networks. (a) The default student network uses the pre-trained COCO and Human3.6 weights. (b) The trained student network exploits the soft and hard pseudo labels obtained from the teacher network. The left side shows the 2D/3D visualization results at 1x scale, and the right side shows the 2D/3D visualization results at 12x scale.}
	\label{figure:qual-results-2}
\end{figure}

\begin{table}[t]
	\begin{center}
		\begin{tabular}{|P{2.0cm}|P{1.4cm}|P{1.4cm}|P{0.8cm}|P{2cm}|P{2.0cm}|P{2.0cm}|}
			\hline
			Network & \#Params & GFLOPs & Scale       & Box-AP         & Keypoint2D-AP  & Keypoint3D-MPJPE \\\hline
			        &          &        & \textbf{1x} & \textbf{53.81} & \textbf{57.78} & \textbf{134.88}  \\\cline{4-7}
			Teacher &  250.1M  & 1048.8 & 8x          & 39.03          & 29.28          & 170.25           \\\cline{4-7}
			        &          &        & 10x         & 31.90          & 18.60          & 203.89           \\\cline{4-7}
			        &          &        & 12x         & 24.38          & 8.89           & 260.83           \\\hline
			\Xhline{2\arrayrulewidth}
			        &          &        & \textbf{1x} & \textbf{52.77} & \textbf{46.17} & \textbf{147.17}  \\\cline{4-7}
			Student &  67.9M   &  215.0 & 8x          & 40.21          & 27.02          & 168.10           \\\cline{4-7}
			        &          &        & 10x         & 34.12          & 20.06          & 181.69           \\\cline{4-7}
			        &          &        & 12x         & 29.19          & 14.42          & 194.35           \\\hline
		\end{tabular}
	\end{center}
	\caption{Baseline results on \emph{MVOR+} for teacher and student networks when no training is performed on OR data. Higher AP and lower MPJPE are better. Student and teacher networks are evaluated at original and low-resolution sizes. The aim is to train the student to reach the same performance as the teacher at high resolution (1x).}
	\label{table:baseline}
\end{table}

\begin{table}[t]
	\begin{center}
		\begin{tabular}{|P{2.8cm}|P{0.8cm}|P{2cm}|P{2.5cm}|P{3.5cm}|}
			\hline
			Student Network      & Scale & Box-AP         & Keypoint2D-AP  & Keypoint3D-MPJPE \\\hline
			ORPose\_{fixed}\_1x  & 1x    & \textbf{50.87} & 55.20          & 134.23           \\\hline
			ORPose\_{fixed}\_8x  & 8x    & 49.50          & 53.50          & 137.40           \\\hline
			ORPose\_{fixed}\_10x & 10x   & 49.01          & 51.98          & 137.71           \\\hline
			ORPose\_{fixed}\_12x & 12x   & 48.23          & 49.88          & 138.83           \\\hline
			\Xhline{3\arrayrulewidth}
			                     & 1x    & 50.59          & \textbf{55.80} & \textbf{134.13}  \\\cline{2-5}
			                     & 8x    & 49.57          & 53.31          & 136.45           \\\cline{2-5}
			ORPose\_{all}        & 10x   & 49.25          & 52.12          & 136.95           \\\cline{2-5}
			                     & 12x   & 47.54          & 49.51          & 138.35           \\\hline
		\end{tabular}
	\end{center}
	\caption{Results of our student network evaluated at original size and low resolution images. ORPose\_{fixed}\_\textbf{s}x (\textbf{s}=1,8,10,12) are trained and evaluated at fixed scale. ORPose\_all is a single model trained on random size low resolution and high resolution images, and evaluated on original size images and fixed scale downsampled images.}
	\label{table:results_method}
\end{table}

\begin{table}[t]
	\begin{center}
		\begin{tabular}{|P{3.8cm}|P{0.8cm}|P{1.5cm}|P{2.5cm}|P{3cm}|}
			\hline
			Student Network          & Scale & Box-AP         & Keypoint2D-AP  & Keypoint3D-MPJPE \\\hline
			Single-branch(hard)      & 1x    & 50.61          & 54.73          & 145.77           \\\hline
			Single-branch(soft)      & 1x    & \textbf{51.04} & 54.70          & 134.20           \\\hline
			Single-branch(hard+soft) & 1x    & 50.95          & 55.11          & 152.28           \\\hline
			Double-branch(hard+soft) & 1x    & 50.59          & \textbf{55.80} & \textbf{134.13}  \\\hline
		\end{tabular}
	\end{center}
	\caption{Ablation study on the student network, by comparing to a single branch trained using hard, soft and hard+soft labels. We achieve the best result when using our proposed two-branch design for both 2D and 3D keypoint estimation.}
	\label{table:results_ablation}
\end{table}

\subsection{Results}
Table \ref{table:baseline} shows the results for the teacher and the student networks on \emph{MVOR+}, along with the network parameter complexity, before training on OR data. These networks are initialized from the COCO and Human3.6 pre-trained network weights. We evaluated both on the original image size (1x) and downsampled images of scale 8x, 10x, and 12x. As shown in the Table \ref{table:baseline}, there exists a margin of 11.6\% 2D keypoint AP. Also, the 3D error in the student network is 12.30 mm more compared to the teacher network. When we evaluate these models on the low-resolution images, we observe a strong decrease in the performance, likely because such low-resolution images were not much represented in the training dataset. The low-resolution results of the teacher network are somewhat worse compared to the student network, possibly due to the multi-stage design of the teacher network, where the poor performance of the current stage affects the next stage. The student network is less affected, likely due to its single-stage design.

%We believe the this due to the fact that the distribution of input images on these extreme low-resolution changes considerably which results in such a large decrease in the performance.

Table \ref{table:results_method} shows the results for our student network when trained using the soft and hard pseudo labels for 2D/3D keypoints obtained from the teacher network. We observe improved performance in all the models when trained with the pseudo labels. ORPose\_{all} achieves nearly the same performance compared to the models trained for specific scale low-resolution images. Performance of ORPose\_all on the high-resolution images nearly reaches the teacher network and on the low-resolution images this network performs much better.
This is illustrated in the qualitative results shown in Fig. \ref{result_1} and Fig. \ref{figure:qual-results-2}. Additional qualitative results for other model variants are available in the supplementary material.
%Suggest single network with better training paradign is able to learn the pose details in domain specific scneario for example operating room in our case
%

\subsection{Ablation Study}
To evaluate the effect of soft-labels on the student network, we keep only one branch for 2D/3D keypoint estimation i.e only one heatmap layer and one 2Dto3D network. We train this single branch keypoint-head with only hard labels, only soft labels, and both hard and soft labels. To train for both the hard and soft labels, the 2D losses are computed using the same heatmap layer and 3D losses are computed using the same 2Dto3D network. As shown in Table \ref{table:results_ablation}, we observe that training with the hard labels hurt the 3D keypoint estimation, and training using only the soft labels achieves good overall results. 2D keypoint estimation is however inferior compared to our two-branch design trained for soft and hard losses.

\section{Conclusion}
In this work, we tackle joint 2D/3D pose estimation from monocular RGB images and propose a self-supervised approach to train an end-to-end and easily deployable model for the OR. We use data distillation to exploit non-annotated data and knowledge distillation to benefit from the high-quality predictions of a multi-stage high capacity pose estimation model. Our approach does not require any ground truth poses from the OR and evaluation on the \emph{MVOR+} dataset suggests its effectiveness. We further demonstrate that the proposed network can yield accurate results on low-resolution images, as needed to ensure privacy, even using a downsampling rate of 12x.
\subsubsection{Acknowledgements}
This work was supported by French state funds managed by the ANR within the Investissements d'Avenir program under reference ANR-16-CE33-0009 (DeepSurg). The authors would also like to thank the members of the Interventional Radiology Department at University Hospital of Strasbourg for their help in generating the dataset.
%\bibliographystyle{splncs04}
%\bibliography{references}

\clearpage
\section*{\centering *** Supplementary Material ***}

% \DeclareDocumentEnvironment{env_comparison_2}{O {10} m}
% {   

%     \caption{2D/3D pose visualization of ground truth and proposed student models ORPose\_fixed\_\textbf{s}x (\textbf{s}=1,12) and ORPose\_all evaluated at scale 1x and 12x.  GT-2D and GT-3D are the visualization results from 2D and 3D ground truth keypoints respectively. }
%     \label{fig}
%     \end{figure}
% }
% {}

\newcommand{\compareone}[1]
{
    \begin{env_comparison_0}[#1]{}
    \end{env_comparison_0}  
}

\newcommand{\comparetwo}[1]
{
    \begin{env_comparison_1}[#1]{}
    \end{env_comparison_1}
    
    \begin{env_comparison_2}[#1]{}
    \end{env_comparison_2}    
}

\bigskip

\par

The pictures below show additional qualitative results for the default and trained student network. The default student network is trained using the public COCO and Human3.6 datasets, and the trained student network exploits the soft and hard pseudo labels obtained from the teacher network during training. The left side shows the 2D/3D visualization results at 1x downsampling scale and the right side shows the 2D/3D visualization results at 12x downsampling scale.\\

\medskip
\foreach \n in {20020000051}{\compareone{\n}} 

\clearpage 
\newpage

\bigskip
\par
The pictures below show additional qualitative results for the proposed student networks ORPose\_all and ORPose\_fixed\_\textbf{s}x (\textbf{s}=1,12) evaluated at scales 1x and 12x. GT-2D and GT-3D provide 2D/3D ground-truth pose visualization. 
\bigskip
\foreach \n in {10010000013,10010000016}{\comparetwo{\n}}
% 
%\foreach \n in {10010000013,10010000014,10010000016,20020000051,20020000059,20020000062,20030000073,20030000074,40020000087,40030000003}{\compare{\n}}
%\foreach \n in {10010000016,40030000003}{\compare{\n}}

%\input{paper2324_complete_results.tex}
\end{document}